\documentclass[10pt,twocolumn,letterpaper]{article}
\usepackage{amsmath}
\usepackage{amssymb}
\usepackage[usenames, dvipsnames]{color}
\usepackage{iccv}
\usepackage{enumitem}
\usepackage{epsfig}
\usepackage{graphicx}
\usepackage{subcaption}
\usepackage{times}

\usepackage[breaklinks=true,bookmarks=false]{hyperref}

\iccvfinalcopy % *** Uncomment this line for the final submission

\def\iccvPaperID{****} % *** Enter the ICCV Paper ID here

\begin{document}

%%%%%%%%% TITLE
\title{The Game Imitation: Deep Supervised Convolutional Networks for Quick Video Game AI}

\author{Zhao Chen\\
Department of Physics\\
Stanford University, Stanford, CA 94305\\
{\tt\small zchen89[at]stanford[dot]edu}
\and
Darvin Yi\\
Department of Biomedical Informatics\\
Stanford University, Stanford, CA 94305\\
{\tt\small darvinyi[at]stanford[dot]edu}
}

\maketitle
%\thispagestyle{empty}
%%%%%%%%% ABSTRACT
\begin{abstract}
   We present a vision-only model for gaming AI which uses a late integration deep convolutional network architecture trained in a purely supervised imitation learning context. Although state-of-the-art deep learning models for video game tasks generally rely on more complex methods such as deep-Q learning, we show that a supervised model which requires substantially fewer resources and training time can already perform well at human reaction speeds on the N64 classic game Super Smash Bros. We frame our learning task as a 30-class classification problem, and our CNN model achieves 80\% top-1 and 96\% top-3 validation accuracy. With slight test-time fine-tuning, our model is also competitive during live simulation with the highest-level AI built into the game. We will further show evidence through network visualizations that the network is successfully leveraging temporal information during inference to aid in decision making. Our work demonstrates that supervised CNN models can provide good performance in challenging policy prediction tasks while being significantly simpler and more lightweight than alternatives.
\end{abstract}
%%%%%%%%% BODY TEXT
\section{Introduction}\label{sec:intro}
In 2014, Google DeepMind successfully used convolutional neural networks with deep reinforcement learning to teach a computer to play classic score-based Atari games\cite{atari}. We propose to study a complementary problem - that of teaching a computer to play modern video games purely through imitation learning; i.e. pure behavioral mimicry of human actions viewed through gameplay screenshots. The imitation learning framework provides several advantages over its deep-Q sibling, not the least of which is the fact that we no longer depend on a carefully crafted loss function or an a priori notion of ``game score.'' The gameplay data now also consists of a reliable ground truth that we assume represents close-to-optimal behavior, which should add further robustness at training time.\\ 
\indent The goal of these studies is not to push performance past that of a deep-Q network; while deep-Q methods have already surpassed human performance for some gaming tasks \cite{atari, alphago}, our model performance at test-time is intrinsically limited by the human-generated dataset that our networks train on, as we provide our model with no additional reinforcement signal. However, we will show that our methods provide reasonable performance at a significantly lower cost in terms of hardware and time, and can be implemented with minimal fine-tuning at inference. The same model can also in principle be applied to various games without much additional tuning, and this portability is attractive from an implementation perspective. We thus present our results without any intent of dethroning deep-Q learning in terms of performance, but rather to demonstrate that a significantly more lightweight approach using CNNs can yield reasonable results, which can be of great interest when using these methods in practice. \\ 
\indent More specifically, we propose training a deep convolutional network on the task of playing Super Smash Bros., a classic game for the Nintendo 64 gaming system. We use as input data gameplay frames ($\sim$128x128px images) and as ground truth keyboard input generated from a human player (who will be one of the researchers). Super Smash Bros was chosen not only for its name recognition within the gaming community, but also because it is more complex (with 3D camera angles, random zooms, etc.) when compared to the 2D state-spaces already thoroughly explored by the Atari experiments in \cite{atari}. We show that given a moderately sized labeled data set, a supervised approach will be able to produce a model that can play this complex game stably within human reaction time at the level of a competent human player. However, we also recognize that such a method might not be appropriate for all cases; for comparison purposes, we will also train the same model on a Mario Tennis dataset to investigate what parts of the model are portable cross-game and what limitations such models might face in the wild.\\
\indent Crucial to the success of our model is its ability to reason \textit{temporally} about the dataset and leverage the many rich correlations along the time axis within video game datasets. Our work is thus also relevant to CNN applications in the realm of video and temporal analysis. 
\section{Background}\label{sec:related}
A close sibling to our method is deep-Q reinforcement learning, a method which has been highly successful for a variety of difficult tasks \cite{atari, alphago}. However, reinforcement learning methods are usually slower to converge and also require careful fine-tuning (the creation of an effective reward function, for example), and so these models can be difficult to work with. The problem is compounded when the learning task is framed as a purely vision problem; only having raw pixels as data puts even more stress on the deep-Q networks, and this is even worse for modern games which incorporate complicated 3D graphics and dynamic camera angles.\\
\indent Imitation learning, like reinforcement learning, seeks to determine the best policy for an agent to take given a state space and a set of possible actions. Unlike reinforcement learning, imitation learning gives the agent access to the state and action pairs generated by an expert agent, which is assumed to act according to a close-to-optimal policy. Such methods are attractive for complex problems, since the agent now has reliable information on optimal behavior rather than having to search for this behavior independently. Thus, the main task at hand is to consider how to best leverage this expert information in training our agent. \\
\indent Our method of purely supervised imitation learning, where we ask the agent to directly clone the actions of the expert, does not currently enjoy much status within the machine learning community. Such methods are known to be prone to issues with robustness \cite{rossbagnell}, primarily because the data and labels are no longer distributed I.I.D. due to temporal correlations within the data. Small deviations in agent behavior might lead it into wildly divergent areas of the state space that are not explored by the expert, causing compounding errors in the agent policy. To circumvent this issue, modern imitation learning methods might look towards inverse reinforcement learning (IRL) for inspiration \cite{ganimitation,inversereinforcement}, which allows an agent to directly learn a policy based on an inferred reward function associated with the expert agent. Or they might go to methods that supplement the training dataset incrementally and allow the agent to more freely explore the full extent of the state space \cite{dagger, autoimitation}. Other work, including the well-known AlphaGo model by Google DeepMind \cite{alphago}, uses imitation learning but only as a priming step to help further learning later in a deep-Q network. \\
\indent In our case, we operate under the assumption that our dataset lies in a sufficiently well-controlled space and that a convolutional network is expressive enough for vision tasks, so that a purely supervised model will still yield reasonable results. In a way, our work is a statement on how CNNs produce expressive features that are exceedingly robust on image datasets. We also circumvent issues that might arise with correlation within our training dataset by explicitly enforcing any such correlations at training time through an appropriate architecture. The hope here is that by allowing the supervised model direct access to temporal information, the network will be not be strictly limited to examples it sees in the training dataset, but will also be able to generalize to unfamiliar situations. Our desire is to create a model that is attractively simple in its implementation and yet still can be useful in practice.\\
\indent As a final note, intimately related to our topic is work on video classification\cite{videoclass, videoclass2, actionrecog}, as our input data is essentially a frame within a larger video context. With reference to previous experiments done on large-scale video classification, we will experiment with various CNN video models to evaluate which is optimal for our learning problem. We discuss these methods and their implementations on our data set in section \ref{sec:model}.
\section{Problem Description}
As mentioned in Section \ref{sec:intro}, we propose a purely supervised learning classification for fully vision-based gaming AI. More precisely, our classification pipeline takes as input a concatenation of four temporally sequential gameplay images, each of dimension $(128,128,3)$, making the input a dimension of $(4,128,128,3)$. We then use a convolutional neural network to output 30 class softmax scores, each corresponding to a different possible input command on the Nintendo 64 console. The classes themselves are selected based on the author's expectations of important button combinations within the game; in Super Smash Bros., for example, the combination down+B is an important button combination and hence is encoded as its own separate class. At inference, we not only want to see that our model can correctly predict button presses on a held-out validation dataset, but also want to evaluate the performance of our neural network in a real game setting, and so will use these class scores in a gameplay simulation to send input commands to a game character playing live against a CPU opponent. 
%-------------------------------------------------------------------------
\section{Data Collection and Preprocessing}
\begin{figure}[htb!]
\begin{center}
   \includegraphics[width=1\linewidth]{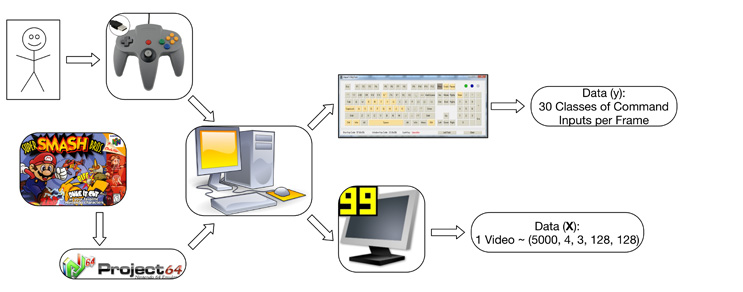}
\end{center}
   \caption{\textbf{Data Collection Pipeline.}} 
\label{fig:pipeline}
\end{figure}
\begin{figure}[htb!]
\begin{center}
   \includegraphics[width=1\linewidth]{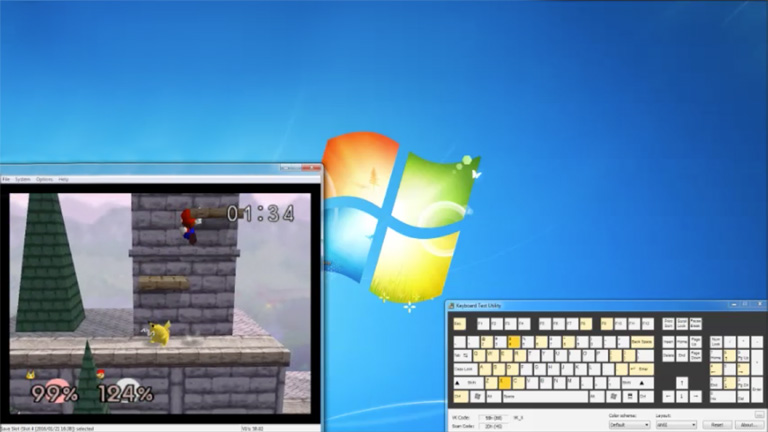}
\end{center}
   \caption{\textbf{Data collection environment.}  Our screen capture setup.  We capture the visual game data together with a visualization of the keys pressed for each frame of gameplay. This simultaneous data collection scheme ensures synchronization between inputs and labels. The keyboard in the bottom right shows currently pushed keys (orange), previously pushed keys (yellow), and inactive keys (white).}
\label{fig:screenshot}
\end{figure}
\begin{figure}[htb!]
	\centering
	\begin{subfigure}[h]{0.3\linewidth}
		\includegraphics[width=\textwidth]{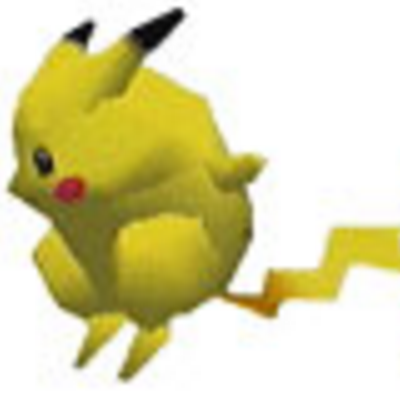}
		\caption{Pikachu}
	\end{subfigure}
	~
	\begin{subfigure}[h]{0.3\linewidth}
		\includegraphics[width=\textwidth]{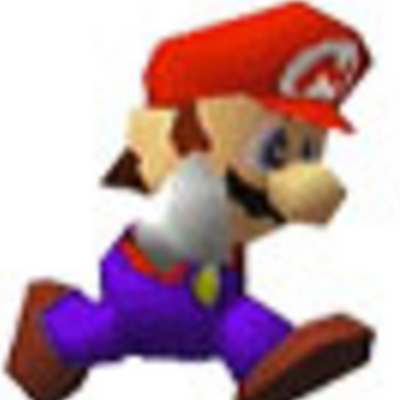}
		\caption{Mario}
	\end{subfigure}
	\caption{\textbf{Character Sprites}}
	\label{fig:sprites}
\end{figure}

The data was collected by the authors via a combination of Nintendo 64 emulation and screen capture tools.  The Nintendo 64 and games (Super Mario Bros and Mario Tennis) were emulated using Project 64 v2.1\cite{pj64}\footnote{For the sake of fair use, we should mention that a purchased copy of Super Smash Bros. and Mario Tennis along with a Nintendo 64 machine are owned by the authors. The emulator software is used purely for academic reasons: i.e. to allow easy interfacing with computer software for data collection.}.  Keyboard Test Utility, which visually registers key pushes on a windows machine, was used in parallel to keep track of the button presses made during the gameplay.  By visualizing keyboard pushes and the gameplay data on the same screen, we can ensure synchronization between the input and label data.  Gameplay footage was captured with the screen capture program Fraps v3.5.99\cite{fraps} at 30 frames per second.  Figure \ref{fig:screenshot} shows a frame of the screen capture data, with the gameplay footage at the bottom left and keyboard visualization at the bottom right.  \\
\indent For the full dataset, 60 5-minute games were played of Super Smash Bros, giving approximately 600,000 frames of gameplay with corresponding key presses. For our later comparison with Mario tennis, 25 single games were played of Mario Tennis, giving approximately 125,000 frames of data. To minimize stress on our eventual models, game parameters were kept constant across all play-throughs.  The player always played as Pikachu against Mario in Super Smash Bros., (see figure \ref{fig:sprites}), and the game stage was also always Hyrule Castle.  These characters were chosen in part because they are a classic matchup in the game, but more importantly their color palettes are very distinct and should allow color information to enter easily into our network.  In Mario Tennis, the player always played Yoshi against Birdo on a clay court. A schematic of the data collection pipeline is shown in Figure \ref{fig:pipeline}.\\
\indent Image preprocessing was kept minimal to keep the training pipeline as end-to-end as possible.  Each frame of the video was downsampled from an original size of $(334, 246)$ (in pixels) to $(128, 128)$.  We downsample our images \textit{not} done by bicubic interpolation, which is a relatively standard image resizing metric, but rather by a nearest neighbor interpolation.  Thus, every pixel in our final downsampled frame will be a pixel from the original image that was close in position to its final downsampled location.  We made this choice because we realized that as these game sprites are somewhat small (they occupy about 1-3\% of the screen as can be seen in figure \ref{fig:screenshot}), bicubic downsampling muted a lot of the colors of these sprites and reduced sharpness of many texture features.  Thus, even though nearest neighbor sampling is known to be a much noisier downsampling metric, we chose to try to preserve the stronger colors of our images.  A mean image (defined as the mean over the training set) was also subtracted from each image beforehand.

%-------------------------------------------------------------------------

\section{Model and Methods}\label{sec:model}
\begin{figure}[htb!]
\begin{center}
\includegraphics[width=\linewidth]{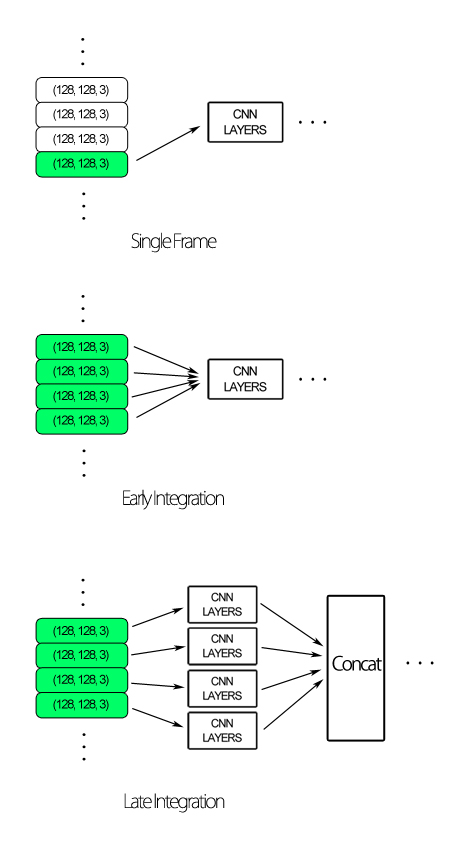}
\end{center}
   \caption{\textbf{Model Selection.}  This figure shows high-level overviews of the three convolutional neural network architectures that we considered.  On top is the single frame CNN, which is just a vanilla image classification CNN.  The middle model is our early integration model as described in section \ref{sec:early}. The bottom model is the late integration model, described in section \ref{sec:late}, and the model we ultimately use.}
\label{fig:integration}
\end{figure}

Three main convolutional neural network architectures were considered as possible model to our frame classification problem: the single frame CNN, a CNN with early integration of video frames, and a CNN with late integration of video frames (schematically shown in figure \ref{fig:integration}).  The foundational CNN architecture for all three of our models is AlexNet \cite{alexnet}.  We will take the standard softmax loss as classification loss in our system, and train with Adam and standard learning hyperparameters. The softmax layer will also be useful at inference by allowing us to treat the final class scores as probabilities from a sampling distribution (see section \ref{sec:testtime}).\\
\indent Below, we describe our three main approaches in detail.  We will then describe how we trained these models and did fine-tuning at test time to optimize performance during live gameplay.
\subsection{Single Frame CNN}\label{sec:single}
For our single frame CNN, we use a vanilla AlexNet structure.  Following the original AlexNet paper \cite{alexnet}, we design our AlexNet to have the following layers:
\begin{enumerate}[topsep=0pt,itemsep=-1ex,partopsep=1ex,parsep=1ex]
\item \texttt{INPUT}: $128 \times 128 \times 3$ \color{black}
\item \color{red} \texttt{CONV7}: $7 \times 7$ size, 96 filters, 2 stride \color{black}
\item \color{ForestGreen} \texttt{ReLU}: $\max(x_i, 0)$ \color{black}
\item \color{Fuchsia} \texttt{NORM}: $x_i = \frac{x_i}{\left(k + (\alpha \sum_j x_j^2) \right)^{\beta}}$ \color{black}
\item \color{blue} \texttt{POOL}: $3 \times 3$ size, 3 stride \color{black}
\item \color{red} \texttt{CONV5}: $5 \times 5$ size, 256 filters, 1 stride \color{black}
\item \color{ForestGreen} \texttt{ReLU}: $\max(x_i, 0)$ \color{black}
\item \color{blue} \texttt{POOL}: $2 \times 2$ size, 2 stride \color{black}
\item \color{red} \texttt{CONV3}: $3 \times 3$ size, 512 filters, 1 stride \color{black}
\item \color{ForestGreen} \texttt{ReLU}: $\max(x_i, 0)$ \color{black}
\item \color{red} \texttt{CONV3}: $3 \times 3$ size, 512 filters, 1 stride\color{black}
\item \color{ForestGreen} \texttt{ReLU}: $\max(x_i, 0)$ \color{black}
\item \color{red} \texttt{CONV3}: $3 \times 3$ size, 512 filters, 1 stride\color{black}
\item \color{ForestGreen} \texttt{ReLU}: $\max(x_i, 0)$ \color{black}
\item \color{blue} \texttt{POOL}: $3 \times 3$ size, 3 stride\color{black}
\item \color{Orange} \texttt{FC}: 4096 Hidden Neurons \color{black}
\item \color{RedViolet} \texttt{DROPOUT}: $p = 0.5$ \color{black}
\item \color{Orange} \texttt{FC}: 4096 Hidden Neurons \color{black}
\item \color{RedViolet} \texttt{DROPOUT}: $p = 0.5$ \color{black}
\item \color{Orange} \texttt{FC}: 30 Output Classes \color{black}
\end{enumerate}
\indent We can see that our single frame CNN turns our supervised learning problem into a simple single image classification problem.  Given any frame, we will learn what would be the most probable button class for that frame.  The validation accuracy as a function of training epoch is shown in figure \ref{fig:validation}.\\
\indent However, such a model ignores temporal information of our data set in our model, which can lead to many of the issues with supervised imitation learning as described in \cite{rossbagnell}.  From an instinctive level, character movement and trajectories should also tie in deeply with a decision-making process for video games; it is clear, for example, that whether players are moving towards each other or away from each other should heavily influence the action taken.  Any reasonable model for this problem, therefore, must use previous frames and character \textit{trajectories} rather than pure current states to better understand higher-order derivative features such as velocity and acceleration. 

\subsection{Early Integration CNN}\label{sec:early}
\indent The most na\"{i}ve incorporation of temporal information would be to concatenate all frames at the onset of the network, forming an input of $128 \times 128 \times 3F$ where $F$ is the number of frames.  In our case, we pick $F=4$ for the remainder of this paper, and so the last dimension of our input here would be 12. This is exactly the idea behind an early integration CNN.  We will use the same layers 2-20 proposed in section \ref{sec:single}, but our first layer will be
\begin{enumerate}[topsep=0pt,itemsep=-1ex,partopsep=1ex,parsep=1ex]
\item \texttt{INPUT}: $128 \times 128 \times 12$ \color{black}
\end{enumerate}
\indent We choose our frames to be $t = \frac{1}{6}$s apart, which gives one input to our model a temporal resolution of 0.5s. Human reaction time is on the order of $0.25s$, which means this temporal resolution provides our network with approximately twice as much information as the human reaction limit. We do not further tune this hyperparameter during training.\\
\indent The main problem with the early integration architecture lies within the fact that in an early integration CNN, only the first convolutional layer has access to the temporal information in a separable form.  Once the input has passed through the first layer, all temporal information is now entangled within the first layer activations. The subsequent layers then do not have direct access to the temporal information, diminishing their ability to learn temporal features.  To circumvent this issue, we turn to the late integration CNN architecture.

\subsection{Late Integration CNN}\label{sec:late}
\begin{figure}[htb!]
\begin{center}
   \includegraphics[width=1\linewidth]{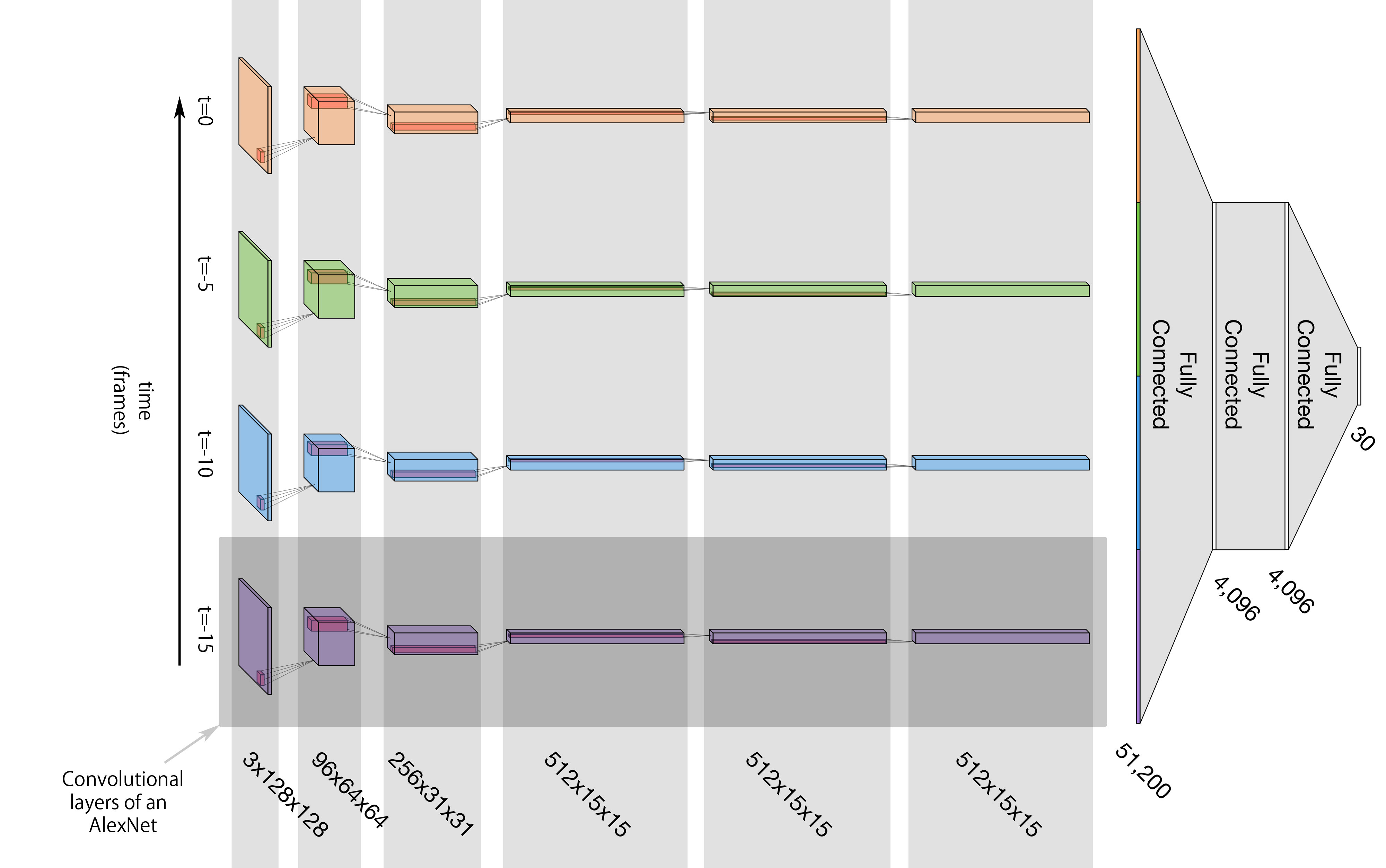}
\end{center}
   \caption{\textbf{Late Integration.}  This figure shows our implementation of a late integration convolutional neural network as described in section \ref{sec:late}.}
\label{fig:lateIntegration}
\end{figure}

\indent Similar to section \ref{sec:early}, we will still consider four time points of image data, each frames in a temporal sequence separated by $\frac{1}{6}$s.  However, rather than concatenating the input data before the first layer, we will send each frame through its own independent CNN layers with untied weights before merging the activations at the end through a series of dense layers.  Each frame will be separately processed by the AlexNet architecture defined in layers 2-15 from section \ref{sec:single}. The dense layers which consolidate the activations from the independent CNN branches will have the same architecture as layers 16-20 in the architecture presented in section \ref{sec:early}. The final model architecture is schematically reproduced in figure \ref{fig:lateIntegration}.\\ \ref{fig:lateIntegration}.
\subsection{Model Training}
We perform training on a NVIDIA GTX 870M GPU (a rather old GPU chip with only 4gb memory) and primarily use Lasagne/Theano to build our network.\cite{theano1, theano2, lasagne} We run hyperparameter search on learning rate and $L_2$ regularization penalty with a small number $(<10)$ of trials. We find that the model training efficacy is generally insensitive to regularization but is very dependent on learning rate, and we settle on the following training parameters: 
\begin{itemize}[topsep=0pt,itemsep=-1ex,partopsep=1ex,parsep=1ex]
\item Learning Rate: 1e-4
\item Learning Rate Annealing: 0.95 every 5000 iterations
\item $L_2$ Regularization Penalty: 1e-7
\item Update Rule: Adam
\item Batch Size: 25 (maximal size on late-integration model given our GPU memory)
\item Training Time: 2 epochs over 2 days.
\end{itemize}
\indent Note that this is significantly less training on much less data when compared to deepーQ learning methods like in \cite{atari}, which used 50 million frames of gameplay.\\
\indent　Batches were randomized across the entire training data set (a single batch may consist of sequences of frames from different gameplay videos). We also do not employ any data augmentation; since all our data was generated in the limited confines of a game emulator, we expect the data set to be relatively robust and for augmentation to have limited effect. 
\subsection{Test-Time Tuning for Live Simulations}\label{sec:testtime}
\begin{figure}[htb!]
\begin{center}
   \includegraphics[width=1\linewidth]{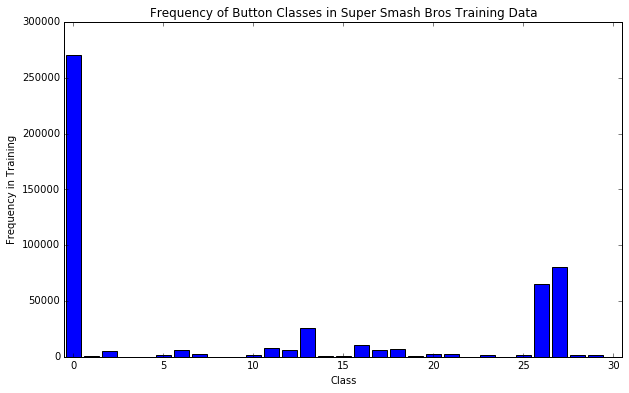}
\end{center}
   \caption{\textbf{Training Data Histogram.}  Histogram of the frequency of all 30 classes in our full training data set.}
\label{fig:histogram}
\end{figure}

At test time, taking the pure maximum softmax class score as the next input for a live simulator does not immediately produce reasonable test-time behavior; the reason for this becomes evident if we look at the histogram in figure \ref{fig:histogram}, which shows the frequency of occurrences of each class in the training data. The dataset is very unbalanced, and classes 0 (None), 26 (Left), and 27 (Right) are by a large margin the most represented in the training set, and as such tend to have overinflated softmax scores at inference. One way to compensate for class unbalance is to train with batch selection or a weighted loss function, but we find that a simple post-processing step that adds scalar weights to class scores already produces impressive test-time performance, which is attractive in that it saves us from retraining the network with additional hyperparameter search. \\
\indent Our solution is to bias each softmax class score by a simple multiplier $b_{class}$, which is chosen in order to ensure that each class is expressed sufficiently during inference. This can be done automatically by selecting $b_{class}$ to balance the confusion matrix (see figure \ref{fig:confusion}) and ensure that the false positive rate is approximately equal across classes. In our case, for example, balancing classes in post-processing suppresses scores for classes 26 and 27 (which induce many misclassifications) to approximately $10\%$ of their original value. Given that the number of classes is low, some further fine-tuning based on prior knowledge of the game reward structure can yield additional performance gains (for example, preferencing crucial classes that are useful for recovering from dire situations). \\
\indent In addition, because our top-3 validation score (see Section \ref{sec:results}) exceeds 96\%, at test-time instead of taking the maximum softmax score as our output class, we take the top 3 class scores for each forward pass and sample from them as a distribution. This allows us to inject a richer set of behaviors into our model and reflects our belief that human behavior in video games is not completely deterministic, and that multiple actions may be appropriate in some circumstances. We also find that this added complexity prevents our model from ``getting stuck'' when presented with some states for which more than one action may have substantial softmax score, but for which only one action may allow the game to continue. \\ 
%------------------------------------------------------------------------
\section{Results and Discussion}\label{sec:results}
Our major results are for Super Smash Bros., and we defer discussion of Mario Tennis until section \ref{sec:tennis}. We use top $N$ validation accuracy as a metric for preliminary model evaluation; we also exhibit a confusion matrix heat map on a test set to visualize specific problem areas with our final classifier. Last, we quantify performance in live games of our neural network player as a crucial evaluation metric. We also provide videos of full simulations and visualizations run on Super Smash Bros. and Mario Tennis, which can be found in section \ref{sec:videos}. Although these videos are purely qualitative, they are convincing demonstrations of our network performance. 
\subsection{Validation Accuracy}
\begin{figure}[htb!]
\begin{center}
   \includegraphics[width=1\linewidth]{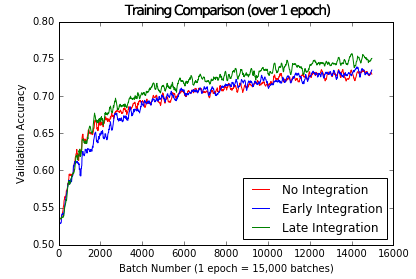}
\end{center}
   \caption{\textbf{Validation Accuracy of our Three Models.}  This figure shows the validation accuracy of all three time-integration models investigated in section \ref{sec:model}.  For the purposes of this comparison, we trained all three models for 1 epoch (15000 batches of 25 frames each).}
\label{fig:validation}
\end{figure}

\indent For comparison of the three main models we described in section \ref{sec:model}, we recorded top-1 validation accuracies over one epoch of training, and the results are shown in figure \ref{fig:validation}. Validation scores are based on a held-out set of full gameplay videos to ensure the validation data is uncorrelated with the training data. We can see that the late integration model performance does begin to exceed that of the other proposed models after a few thousand batches; the difference may seem moderate at best ($\sim 1\%$) in terms of validation accuracy, but it can be quite significant in live simulation performance (see section \ref{sec:videos}), and the late-integration model also produces convincing network visualizations (see section \ref{sec:saliency}) that explicitly demonstrate the network is learning to properly parse temporal information.\\
\indent A confusion matrix for our classification problem is shown below in figure \ref{fig:confusion}.  The confusion matrix heatmap shown on the right is normalized to class frequency; we see clearly that classes 0, 26, and 27 induce by far the most misclassifications. This is very consistent with the the issue with lopsided training data described earlier in section \ref{sec:testtime}, demonstrating that some test-time tweaking (which in our case comes in the form of class-score multipliers) is well-justified.
\begin{figure}[htb!]
	\centering
	\begin{subfigure}[h]{0.45\linewidth}
		\includegraphics[width=\textwidth]{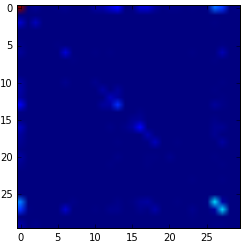}
		\caption{Raw Confusion Matrix}
	\end{subfigure}
	~
	\begin{subfigure}[h]{0.45\linewidth}
		\includegraphics[width=\textwidth]{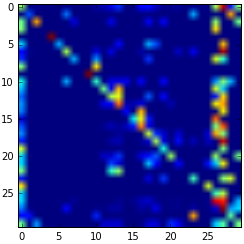}
		\caption{Normalized by Label Frequency}
	\end{subfigure}
	\caption{\textbf{Confusion Matrix Heatmap}}
	\label{fig:confusion}
\end{figure}
After 2 epochs of training each, we arrive at the validation scores reported in table \ref{table:valacc}. Top 3 accuracy for Super Smash Bros. exceeds $96\%$, which shows that our model is learning correct behavior but that this behavior might not produce the maximal softmax score. Note that we only performed minimal hyperparameter search, and so we expect these validation scores to be essentially unbiased. 
\begin{table}[h]
\footnotesize
\centering
\begin{tabular}{c c c c}
 Game & Top 1 Acc & Top 3 Acc & Top 5 Acc\\
\hline
Super Smash Bros.& 79.9\% & 96.2\% & 98.7\%\\
Mario Tennis& 75.2\% & 96.0 \% & 99.5\%
\\
\end{tabular}
\caption{Validation Accuracy After 2 Epochs Training}
\label{table:valacc}
\end{table}
\subsection{Performance on Live Gameplay}\label{sec:videos}
\begin{figure}[htb!]
\begin{center}
   \includegraphics[width=1\linewidth]{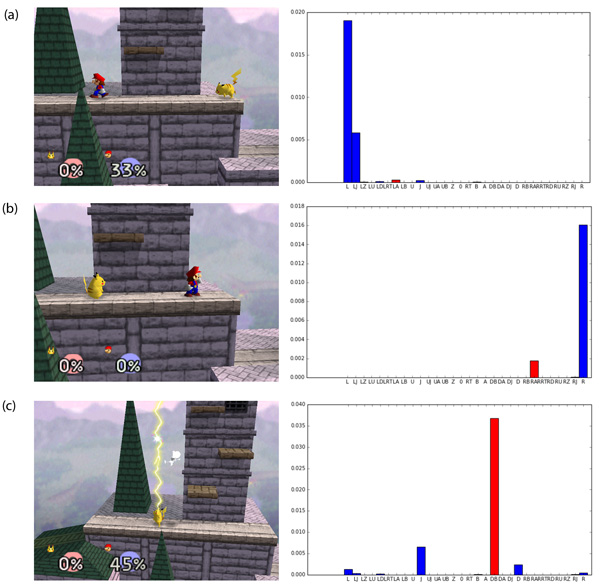}
\end{center}
   \caption{\textbf{Class Score Visualization}  Game footage (left) is juxtaposed with class scores (right) at that moment in time. Blue bars in the histogram code for passive actions (movement), while red for aggressive actions (attacks). These frames exhibit strong class scores in one category, namely (a) move left, (b) move right, (c) down+B attack. These class scores suggest that the network has properly learned character tracking information and is sensitive to the relative positioning of the two character sprites.}
\label{fig:class_scores}
\end{figure}

We find that our model runs at test-time with a latency of 300ms (on our mediocre GPU). This is only slightly below the average human reaction time ($\sim$ 250ms) and so we can expect realistic reaction speeds when running this network in simulation. \\ 
\indent To evaluate our model's practical performance, we pit our fully CNN-controlled Pikachu against the pre-packaged Mario game AI. We run 10 games against a level 9 (highest level) Mario AI and 10 games against each a level 3 and level 6 Mario AI and record damage dealt in each game. The results are summarized in Table \ref{table:matches}. Margins of error are 95\% confidence intervals assuming Gaussian data distributions. We note that the CNN Pikachu handily defeats a level 3 CPU, is just barely outside the margin of error for being better than a level 6 CPU, and is quite competitive with the level 9 CPU (they are statistically tied). 
\begin{table}[h]
\footnotesize
\centering
\begin{tabular}{c c c}
 &Average Damage Dealt&Average Damage Dealt\\
Mario CPU Level& by Pikachu per Game & by Mario per Game\\
\hline
3 & $116.6\%\pm 29.3\%$& $66.4\%\pm 12.6\%$\\
6 &  $96.4\% \pm 13.8\%$& $80.5\%\pm 18.7\%$\\
9 & $77.2\%\pm 18.3\%$ & $71.6\%\pm 24.7\%$
\\
\end{tabular}
\caption{CNN Pikachu Performance Against Mario AI}
\label{table:matches}
\end{table}
One important point to note is that our training data is generated through matches against a level 9 CPU opponent. The fact that our CNN Pikachu can also perform well against other level CPU opponents with their own set of behaviors is encouraging, as it supports our statement that our CNN model generalizes well. \\
\indent We further find in live simulations that our CNN Pikachu successfully learned how to track Mario. That is, the our CNN output is sensitive to what direction the Mario character model is relative to Pikachu, and Pikachu is inclined to move towards Mario. This is illustrated in Figure \ref{fig:class_scores}, which shows the class softmax score distribution. We see that Mario's relative position to Pikachu causes strong class predictions when Mario is (a) to Pikachu's left, (b) to Pikachu's right, and (c) above Pikachu, respectively. In case (a), we see a strong class score for left movement and left jump. In case (b), we see a strong class score for right movement. In case (c), we see a strong class score for the down+B attack move, which is a move that most couples to the space above Pikachu. All three of these trends are consistent with conventional gameplay wisdom, and suggests that our CNN Pikachu can successfully track his opponent's position on the game screen.\\
\indent This analysis is of course a bit disingenuous, as our algorithm takes as input four frames in temporal sequence, not just one. We will take a closer look at how our network is coupling to all the four frames of input in section \ref{sec:saliency}.
\subsection{Gameplay Saliency Analysis}\label{sec:saliency}
\begin{figure*}[htb!]
\begin{center}
   \includegraphics[width=0.85\linewidth]{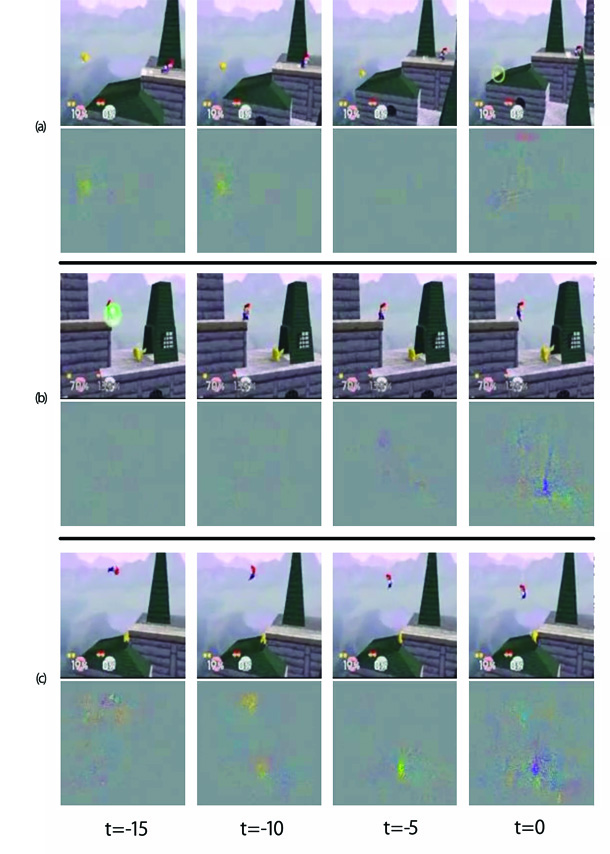}
\end{center}
   \caption{\textbf{Saliency map analysis of live gameplay.}  Saliency map for all four frame inputs during live gameplay simulation. Our CNN model clearly reacts differently to frames at different points in the temporal sequence.}
\label{fig:saliency}
\end{figure*}

\begin{figure*}[htb!]
\begin{center}
   \includegraphics[width=0.85\linewidth]{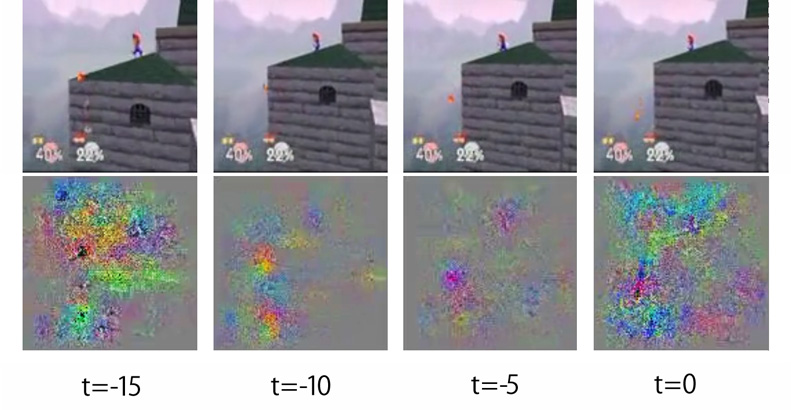}
\end{center}
   \caption{\textbf{Edge sensitivity.} The network is sometimes able to detect edges in the image background when these features become relevant (see the $t=0$ saliency map).}
\label{fig:saliencyedge}
\end{figure*}

We choose saliency maps (as introduced in \cite{saliency}) to visualize our network and qualitatively evaluate how it parses temporal information within our dataset. In essence, a saliency map calculates which pixels produce positive gradient in a particular class softmax score; in our case, we produce saliency maps at each frame during live gameplay simulation for the class that the network predicts (since this is a simulation, we have no access to ground truth labels). Our saliency maps are evaluated in color, unlike traditional saliency maps which are black and white. Gradient values are clipped at $\pm$exp[-9] and the resultant interval is rescaled to lie in $[0,256]$. \\
\indent These saliency maps expose some interesting behavior within our networks. In figure \ref{fig:saliency}, we can see full sets of saliency maps for selected frames. Crucially, we see a clear specialization of the network in processing the different frames: some situations (e.g. figure \ref{fig:saliency}a) couple more strongly to the earlier temporal information, while some situations (e.g. figure \ref{fig:saliency}b) couple more strongly to more recent temporal information. In general, these couplings are consistent with conventional wisdom in the context of the game: scenarios in which the player is flying away and for which player trajectory is important in influencing agent behavior seem to couple strongly to earlier frames, while scenarios in which the player encounters a nearby danger and must react quickly couple strongly to more recent ones. In a sense, earlier temporal information acts partially as a proxy for information on temporal derivatives, allowing the network to be cognizant of the velocity and acceleration of the player. \\
\indent Figure \ref{fig:saliency}c perhaps shows the most interesting behavior, in that all four frames show relatively localized saliency signals, but the pixels which light up are markedly different. The more recent frames couple strongly to Pikachu, while the more temporally distant frames couple more strongly to Mario, with the $t=-10$ frame coupling to both. We interpret this as an indication that the network is successfully processing not only the spatial positioning of the characters, but deriving useful features from character \textit{trajectories}.\\
\indent The network also does not seem limited to processing character information, but also couples to features in the background behind the characters. In figure \ref{fig:saliencyedge}, we can see some clear edge features present in the background show up in the $t=0$ saliency map, which influences Pikachu to execute an Up+B move in an attempt to jump back onto the map. 

\subsection{Performance on Mario Tennis}\label{sec:tennis}
\begin{figure}[htb!]
\begin{center}
   \includegraphics[width=1\linewidth]{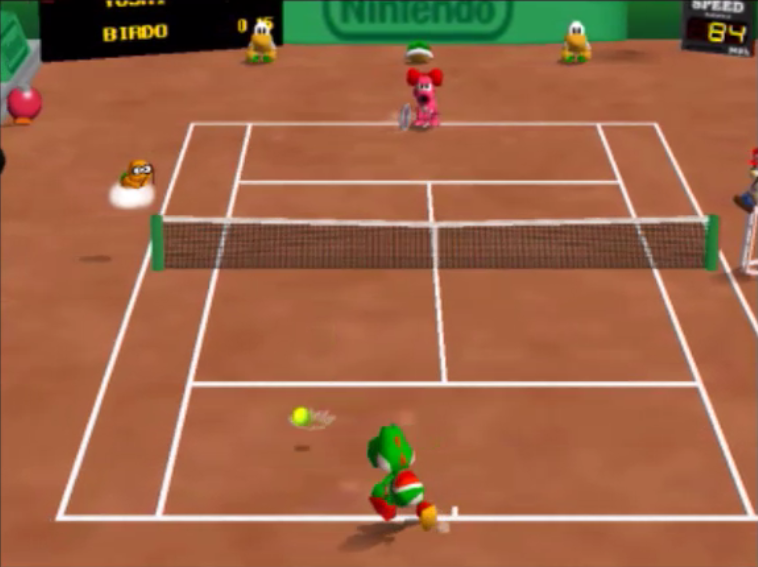}
\end{center}
   \caption{\textbf{Automatic Play of Mario Tennis.}  Yoshi tracks and runs towards the ball in Mario Tennis live gameplay.}
\label{fig:tennis}
\end{figure}

We also train the same late-integration model on a Mario Tennis dataset generated in the same fashion. The model yielded (as shown in table \ref{table:valacc}) top-1, top-3, and top-5 validation accuracies of 75.2\%, 96.0\%, and 99.5\%, respectively. These numbers may seem comparable to our accuracies for the Super Smash Bros. model, but the Mario Tennis labels only consisted of ten classes, so these numbers are slightly weaker. Still, in live gameplay simulation with the same tweaks as described in section \ref{sec:testtime}, we find that a CNN controlled Yoshi (see figure \ref{fig:tennis}) can track the ball successfully and exhibits behavior that is reasonable and learned from our gameplay dataset. However, the live gameplay (see videos linked in \ref{sec:videos}) is significantly less convincing when compared to that of our Super Smash Bros. CNN, and we discuss why this might be the case in section \ref{sec:limitations}.
\subsection{Model Limitations}\label{sec:limitations}
We have already shown that our model can play at close to human reaction times and effectively against difficult CPU AI. It does so without any knowledge of game objectives, and can work across titles with different game objectives and styles of play. And it does so in a supervised learning setting, which allows our models to train faster on fewer data compared to deep-Q learning. However, the degraded performance on the Mario Tennis dataset suggests that there may be some limitations to our model, and we try to list those limitations here:
\begin{itemize}[topsep=0pt,itemsep=-1ex,partopsep=1ex,parsep=1ex]
\item \textbf{Long-term memory} Because we have hard-coded the temporal separation between subsequent frames input into our model (i.e. $\frac{1}{6}$s), the temporal information our model has is limited to $\frac{1}{2}$s into the past. Temporal information beyond that might be important, especially in games like Mario Tennis where consistent tracking of the ball over longer periods is necessary. 
\item \textbf{Allowable error} Intrinsic to our model is the idea that the output is modeled by a probability distribution given by our softmax scores. This assumes that there may be multiple allowable button choices given a game state, and in games where button uncertainty might lead to fatal mistakes such an assumption might not be optimal. This can certainly be the case in games where long-term tracking of game entities is crucial to performance, as such tracking would rely on consistency in button predictions.
\item \textbf{Color specificity} Because it is clear that our model is coupling strongly to colors (see section \ref{sec:saliency}), the current model does not generalize easily to different maps and characters due to the differing color palettes. It may be possible to correct this generalizability issue by incorporating color offsets during data preprocessing and making good use of color augmentations during training, but such methods were not employed yet for these studies.
\end{itemize}
However, our model has still shown to be interestingly robust for games like Super Smash Bros., where actions are relatively localized in time (i.e. do not need information from the far past) and some prediction uncertainty is allowable if not encouraged. The CNN also lends a substantial amount of generalizability to the model; even though plenty of states during live simulation are not exactly present within the training set, the CNN learns relevant features (relative character positions, detection of map edges) that allows the model to reason about unfamiliar states. 
\subsection{Video Demonstrations}\label{sec:videos}
There is perhaps no clearer qualitative demonstration of our model's performance than a full live gameplay simulation. You can view a full Youtube playlist of some videos relating to this project at:\\
\noindent \url{https://www.youtube.com/watch?v=c-6XcmM-MSk&list=PLegUCwsQzmnUpPwVv8ygMa19zNnDgJ6OC}\\
%\noindent \url{MSk\&list=PLegUCwsQzmnUpPwVv8ygMa19zNnDgJ6OC}\\
\indent In total, there are 6 videos in the play list:
\begin{enumerate}[topsep=0pt,itemsep=-1ex,partopsep=1ex,parsep=1ex]
\item Super Smash Bros. AI Demo 1
\item Super Smash Bros. AI Demo 2
\item Super Smash Bros. AI Demo 3
\item Mario Tennis AI Demo
\item Super Smash Bros. Data Acquisition Time Lapse
\item Super Smash Bros. Single Frame AI Demo (Vanilla Alex-Net)
\end{enumerate}
We can see from the vanilla AlexNet implementation of our model that the behavior of the agent is less rich compared to the late integration models (AI Demos 1-3). The Mario Tennis demo also shows reasonable test-time behavior and signs that the network is beginning to learn tracking, but the CNN-controlled agent is not as competitive with its CPU-controlled opponent.\\ 
\indent In addition to the above, also of interest may be some specific videos which feature additional visualizations that provide added insight into network behavior:
\begin{enumerate}
\item Live demo with full view of simulation environment, including keys pushed by the network: \url{https://youtu.be/2A34d2O_Zb0}
\item Saliency map visualization for full demo: \url{https://youtu.be/f02OvO46fBo}
\item Saliency map visualization in slow-motion: \url{https://youtu.be/rExHINi5-IY}
\end{enumerate}
%------------------------------------------------------------------------
\section{Conclusions}
We have shown that when re-framing a complicated video game learning problem in terms of a simpler supervised learning framework, we can achieve good results on complex games with much less training time and training data. For Super Smash Bros in particular, we achieve AI behavior that can compete with the most advanced CPU AI in the game with only access to game visuals, and with very little training (only 2 epochs of training over 2 days). \\ 
\indent Also, despite our complicated state space, our model appears to generalize well despite having a rather finite training set. We attribute this to both the expressiveness of CNN features in image analysis, as well as the power of the late-integration model in parsing temporal data. Both of these effects help us mitigate issues with covariate shift and model robustness that are well-documented for supervised imitation learning models. \\
\indent We have thus presented a model that performs well on complex video game tasks while being lightweight and straightforward to train. We believe that the simplicity of our model makes its implementation especially appealing, and hope that our analysis of our model's capacity to reason spatially and temporally about our data lends insight into the generalizing power of deep CNNs on difficult vision problems.
%------------------------------------------------------------------------
\section{Acknowledgements}
We would like to thank the course staff of Stanford's CS231n: Convolutional Neural Networks for Visual Recognition, for all their valuable advice and for giving us the impetus to start this project. 

{\small
\bibliographystyle{ieee}
\bibliography{refnew}
}

\end{document}